\documentclass[manuscript=article]{achemso}

\usepackage{amsmath}

\usepackage[utf8]{inputenc} 
\usepackage[T1]{fontenc}    
\usepackage{hyperref}       
\usepackage{url}            
\usepackage{booktabs}       
\usepackage{amsfonts}       
\usepackage{nicefrac}       
\usepackage{microtype}      
\usepackage[table]{xcolor}         
\usepackage{graphicx}
\usepackage{multirow}
\usepackage{float}
\usepackage{amssymb}
\usepackage{array}
\usepackage{subcaption}
\usepackage{mathtools} 
\usepackage{nicematrix}
\usepackage{tablefootnote}
\usepackage{natbib}
\usepackage{subcaption}
\usepackage{standalone}
\usepackage{amsmath}
\usepackage{siunitx}
\usepackage{algorithm}
\usepackage{algpseudocode}
\usepackage{import}
\usepackage[draft]{minted}
\usepackage{multirow}
\usepackage{siunitx}
\usepackage{listings}

\graphicspath{ {./figures/} }

\author{Dule Shu}
\affiliation[MechE]{Department of Mechanical Engineering,
Carnegie Mellon University,
Pittsburgh PA, USA}
\altaffiliation{Contributed equally to this work}
\author{Zijie Li}
\affiliation[MechE]{Department of Mechanical Engineering,
Carnegie Mellon University,
Pittsburgh PA, USA}
\altaffiliation{Contributed equally to this work}
\author{Amir Barati Farimani}
\affiliation[MechE]{Department of Mechanical Engineering,
Carnegie Mellon University,
Pittsburgh PA, USA}
\alsoaffiliation[ML]{Machine Learning Department,
Carnegie Mellon University,
Pittsburgh PA, USA}
\alsoaffiliation[Chem]{Department of Chemical Engineering,
Carnegie Mellon University,
Pittsburgh PA, USA}
\email{barati@cmu.edu}

\title{A Physics-informed Diffusion Model for \\ High-fidelity Flow Field Reconstruction}

\keywords{Denoising Diffusion Probabilistic Models, Computational Fluid Dynamics, Super-resolution}
\begin{document}

\begin{abstract}
Machine learning models are gaining increasing popularity in the domain of fluid dynamics for their potential to accelerate the production of high-fidelity computational fluid dynamics data. However, many recently proposed machine learning models for high-fidelity data reconstruction 
require low-fidelity data for model training. Such requirement restrains the application performance of these models, since their data reconstruction accuracy would drop significantly if the low-fidelity input data used in model test has a large deviation from the training data. To overcome this restraint, we propose a diffusion model which only uses high-fidelity data at training. With different configurations, our model is able to reconstruct high-fidelity data from either a regular low-fidelity sample or a sparsely measured sample, and is also able to gain an accuracy increase by using physics-informed conditioning information from a known partial differential equation when that is available.
Experimental results demonstrate that our model can produce accurate reconstruction results for 2d turbulent flows based on different input sources without retraining.

\end{abstract}

\section{Introduction}
High-fidelity numerical simulation of fluid dynamics offers valuable information on how engineering systems interact with fluid flows, and is therefore of great interest to engineering design and related fields. Numerical methods for high-fidelity computational fluid dynamics (CFD) simulation, such as the Direct Numerical Simulation (DNS)\cite{moin1998direct}, often require numerically solving the Navier-Stokes equations at a fine scale in both space and time. Such methods have a high computational expense, especially for simulations with high Reynolds numbers. Various fluid modeling techniques have been developed to reduce the computational cost of DNS, including the Reynolds Averaged Navier-Stokes (RANS) modeling \cite{k-eps-launder, k-omega-wilcox}, the Large Eddy Simulations (LES)\cite{smagorinsky1963les, les-combustion}, hybrid RANS-LES models\cite{shur2008hybrid, davidson2006hybrid, walters2013investigation, jakirlic2010numerical, rona2020hybrid, li2010hybrid}, 
functional sub-grid models which determine an effective eddy viscosity by wave number and an \textit{a priori} specified mixing length\cite{maulik2019sub, tabeling2002two, boffetta2012two, pearson2017evaluation, pearson2018log}, and the Explicit LES closure models\cite{bardina1980improved, stolz1999approximate, layton2003simple, mathew2003explicit, san2018generalized}. 
These techniques increase the computational efficiency of the CFD simulation at the expense of reducing the fidelity of the simulation results. 
The underlying conflict between computational complexity and simulation fidelity motivates the study of using machine learning for surrogate modeling to accelerate CFD simulation. Machine learning-based surrogate models cover a wide range of data representations and problem formulations for CFD data, including learned CFD solvers based on Lagrangian representation \cite{ummenhofer2019lagrangian, li2022graph, gns-fluid, random-forrest-fluid} or Eulerian representation \cite{zhumekenov2019fourier, wang-flow-pred-kdd, mesh-graph-net, learned-coarse-turbulence, mp-pde-solver, latent-physics, Thuerey-2020-airfoil, li2022transformer, pant2021deep, barati2017deep}. In this work, we aim to develop a machine learning model which reconstructs high-fidelity CFD data from low fidelity input. Since low-fidelity data requires less computational resources to generate, 
by using a machine learning model to reconstruct high-fidelity data from the numerically-solved low-fidelity one, we can mitigate the conflicts between computational cost and simulation accuracy, and increase the cost-effectiveness of CFD simulations. 


Inspired by the various research progress in deep learning for image super-resolution
, such as the progressive GAN training approach \cite{karras2017progressive, park2019semantic,wang2018fully}, the transformer-based approach\cite{lu2022transformer, yang2020learning, cao2021video}, and the diffusion model-based approach\cite{saharia2022image, ho2022cascaded, li2022srdiff}, several neural network models have been proposed to increase the resolution and reconstruct under-resolved CFD simulations.
Pant \textit{et al.}\cite{pant2020deep} proposed a CNN-based U-Net model to reconstruct turbulent DNS data from the filtered data and achieved state-of-the-art results in terms of Peak Signal to Noise Ratio (PSNR) and Structural Similarity Index Metric (SSIM). 
With a design using skip-connections and multi-scale filters, Fukami \textit{et al.}\cite{fukami2021machine, fukami2019super}
proposed a convolutional neural network-based model to reconstruct high-resolution flow field data from low-resolution data in both space and time domains.
A multi-scale enhanced super-resolution generative adversarial network with a physics-based loss function is proposed by Yousif \textit{et al.}\cite{yousif2021high} to reconstruct high-fidelity turbulent flow with minimal use of training data.
To address the case where low and high-resolution flow fields are not matched, Kim \textit{et al.}\cite{kim2021unsupervised} proposed a CycleGAN\cite{zhu2017unpaired}-based model trained with unpaired turbulence data for super-resolution. 
The effectiveness of generative neural networks in CFD data super-resolution is not limited to the 2d domain but also 3d volumetric flow data\cite{3d-fluid-recons, xie2018tempogan} and particle-based fluids\cite{li2021tpu, tranquil-clouds}.
The learned reconstruction networks can also be coupled with a numerical solver to build effective coarse-grained simulators. Kochkov \textit{et al.}\cite{kochkov2021machine} propose to use a convolutional neural network to correct the error of a simulator that runs on an under-resolved grid. Um \textit{et al.}\cite{um2020solver} generalize the idea of using neural networks to correct the error of numerical PDE solution arising in under-resolved discretization, and incorporate the solver into the training loop by running differentiable solver on the reconstructed results to account for future loss.

The aforementioned deep learning models have yielded promising results in their respective applications and problem settings, yet they share one common limitation: The models are all trained to fit a particular type of under-resolved CFD data as specified by their corresponding training datasets (\textit{e.g.} a specific filter). As a result, if a trained model is used to reconstruct high-fidelity CFD data from low-fidelity input that significantly deviates from the training dataset (\textit{e.g.}, in terms of resolution or a Gaussian blurring process), the accuracy of data reconstruction will drop significantly. In other words, to ensure the best performance, the users will always have to retrain those models when a new set of under-resolved CFD data is given. The dependency on model retraining has significantly restrained the applicability of deep learning tools in high-fidelity CFD data reconstruction. To resolve this issue, we propose a diffusion-based deep learning framework for CFD data super-resolution. In this framework, we reformulate the problem of reconstructing high-fidelity CFD data from low-fidelity input as a problem of data denoising, and use a Denoising Diffusion Probabilistic Model (DDPM)\cite{ho2020denoising} to reconstruct high-fidelity CFD data from noisy input. 
Motivated by the advancement in solving fluid mechanics problems using the Physics-informed neural networks (PINNs) \cite{raissi2019physics, karniadakis2021physics, cai2022physics}, we proposed a procedure to incorporate physics-informed conditioning information in diffusion model training and sampling, which increases the data reconstruction accuracy by making use of the PDE that determines the fluid flow.
Experimental results show that our diffusion model is able to produce comparable results to the state-of-the-art models on the task of high-fidelity CFD data, where it remains accurate in terms of kinetic energy spectrum and PDE residual loss under different input data distribution but without any retraining.

\section{Method}

\subsection{Problem Formulation}
\begin{figure}
    \centering
    \includegraphics[width=0.85\textwidth]{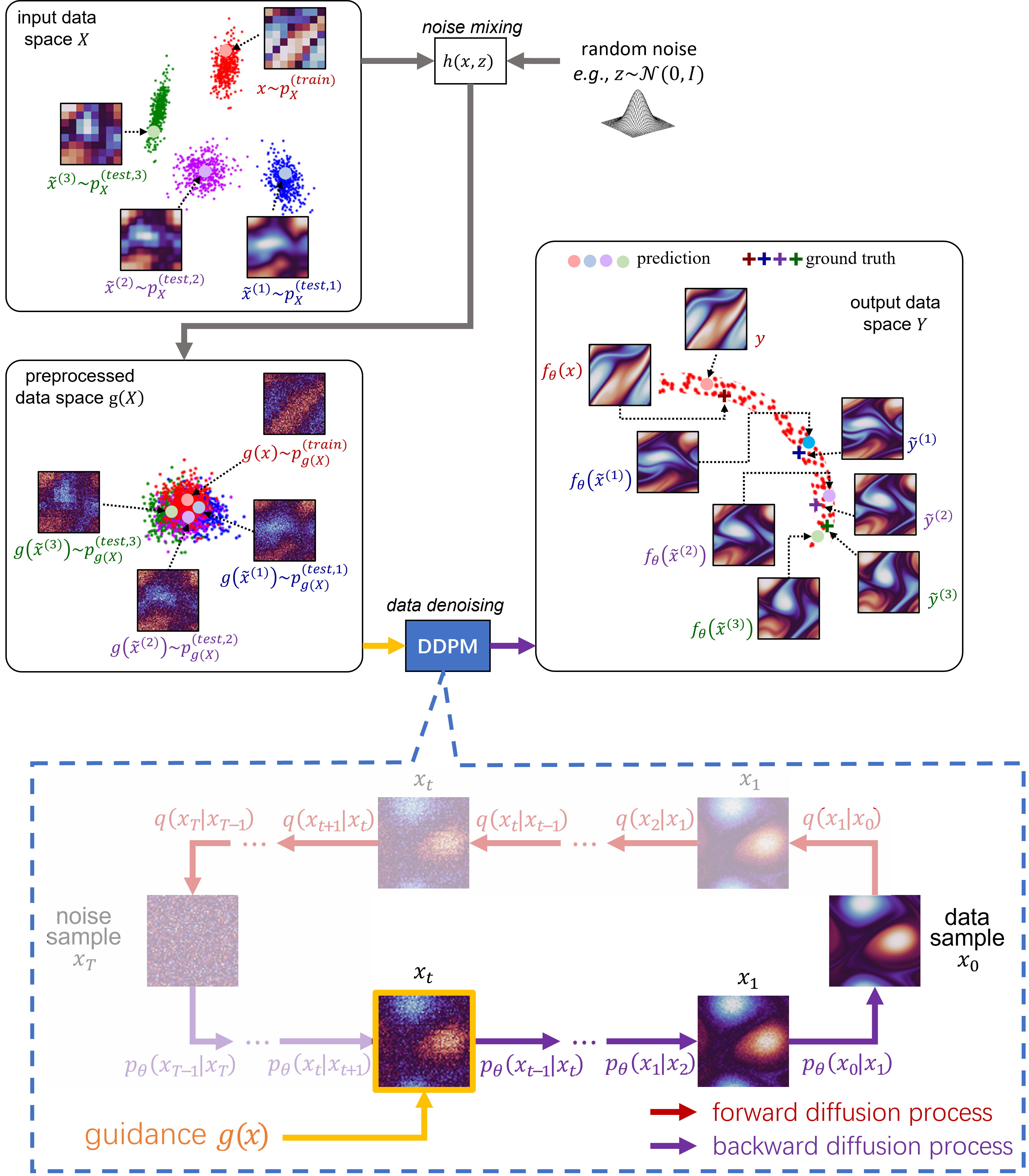}
    \caption{An overview of our proposed framework for high-fidelity CFD data reconstruction using DDPM model. Given input of low-fidelity CFD data samples ($x,\Tilde{x}^{\left(1\right)},\Tilde{x}^{\left(2\right)},\Tilde{x}^{\left(3\right)}$) with various distributions ($p_{X}^{\left(\text{train}\right)}, p_{X}^{\left(\text{test,1}\right)}, p_{X}^{\left(\text{test,2}\right)}, p_{X}^{\left(\text{test,3}\right)}$, respectively), we apply a preprocessing procedure $g:=h(x,z)$ which mixes the input data $x$ with random noise $z$, such that the resulting noisy data samples ($g\left(x\right),g\left(\Tilde{x}^{\left(1\right)}\right),g\left(\Tilde{x}^{\left(2\right)}\right),g\left(\Tilde{x}^{\left(3\right)}\right)$) have much more similar distributions ($p_{g\left(X\right)}^{\left(\text{train}\right)}, p_{g\left(X\right)}^{\left(\text{test,1}\right)}, p_{g\left(X\right)}^{\left(\text{test,2}\right)}, p_{g\left(X\right)}^{\left(\text{test,3}\right)}$, respectively). We then send the noisy data samples to a pretrained DDPM model for data denoising. The output of the DDPM model is a group of noise-free data samples ($f_{\theta}\left(x\right), f_{\theta}\left(\Tilde{x}^{\left(1\right)}\right), f_{\theta}\left(\Tilde{x}^{\left(2\right)}\right), f_{\theta}\left(\Tilde{x}^{\left(3\right)}\right)$) which are close reconstruction of the corresponding ground truth samples ($y,\Tilde{y}^{\left(1\right)},\Tilde{y}^{\left(2\right)},\Tilde{y}^{\left(3\right)}$) of high-fidelity. Inside the DDPM-based data denoising module, we implement a partial backward diffusion process, which is a Markov process starting from a conditioning sample $x_t = g(x)$ and ends at a denoised data sample $x_0=f_{\theta }\left(x\right)$, as shown in the dotted blue box.}
    \label{fig:superresolution-overview}
\end{figure}

Let $f_\theta:X\rightarrow Y$ be a machine learning model which maps data samples from a low-fidelity data domain $X$ to a high-fidelity data domain $Y$. By optimizing model parameters $\theta$ over a training dataset $\left(X^{\left(\text{train}\right)}, Y^{\left(\text{train}\right)}\right)$, we aim to make $f_\theta$ achieve a high data reconstruction accuracy on a test dataset $\left(X^{\left(\text{test}\right)}, Y^{\left(\text{test}\right)}\right)$. 
Let $p_X^{\left(\text{train}\right)}$, $p_X^{\left(\text{test}\right)}$ denote the data distributions of $X^{\left(\text{train}\right)}$ and $X^{\left(\text{test}\right)}$, respectively. Ideally, for a well-trained model to achieve high data reconstruction performance on the test dataset, we need to have $p_X^{\left(\text{train}\right)} \approx p_X^{\left(\text{test}\right)}$.
In practice, however, such condition is generally not satisfied.
If $p_X^{\left(\text{train}\right)}$ and $p_X^{\left(\text{test}\right)}$ have a large difference, the data reconstruction accuracy will drop significantly. 
One idea to mitigate this potential issue is to introduce a preprocessing procedure $g:X\rightarrow X$, such that the preprocessed low-fidelity data sampled from $X^{\left(\text{train}\right)}$ and $X^{\left(\text{test}\right)}$ have similar distributions (\textit{e.g.}, $p_{g(X)}^{\left(\text{train}\right)}\approx p_{g(X)}^{\left(\text{test}\right)}$). 
One way to increase the similarity between $p_{g(X)}^{\left(\text{train}\right)}$ and $p_{g(X)}^{\left(\text{test}\right)}$ is to add Gaussian noise to the low-fidelity data samples, such that both $p_{g(X)}^{\left(\text{train}\right)}$ and $p_{g(X)}^{\left(\text{test}\right)}$ are drawn towards a Gaussian distribution from $p_X^{\left(\text{train}\right)}$ and $p_X^{\left(\text{test}\right)}$, and subsequently become more similar to each other. 
Since our goal is to reconstruct high-fidelity CFD data, after introducing Gaussian noise via the preprocessing procedure $g\left(\cdot\right)$, we need a denoising procedure to  obtain noise-free high-fidelity results. 
Recent works\cite{meng2021sdedit,lugmayr2022repaint,sasaki2021unit,saharia2022palette} have shown successful examples in using a pretrained DDPM model for guided image synthesis and editing. These examples demonstrate the potential of DDPM model in eliminating Gaussian noise from CFD data, and motivate us to use it for our denoising task. 
An overview of our high-fidelity CFD data reconstruction framework with DDPM as the denoising module is shown in Figure \ref{fig:superresolution-overview}.
More details of the DDPM model and how it is used for CFD data reconstruction are provided in the following subsections.

\subsection{Denoising Diffusion Probabilistic Model}
A DDPM model is a generative model that generates data of interest using a Markov chain starting from a sample of standard Gaussian distribution. Let $x_0$ be the data sample to generate, and let $\theta$ be a set of neural network parameters of the DDPM model, the Markov chain can be represented as follows.
\begin{equation}\label{eq:backward-diffusion}
    p_{\theta}\left( x_{0:T} \right):=p\left( x_T \right) \prod_{t=1}^{T} p_{\theta}\left( x_{t-1}|x_{t} \right)
\end{equation}
where $p\left( x_{T} \right):=\mathcal{N}\left( x_T;\textbf{0},\textbf{I}  \right)$ and the probability transition $p_{\theta}\left( x_{t-1}|x_{t} \right)$ is chosen as $p_{\theta}\left( x_{t-1}|x_{t} \right):=\mathcal{N}\left( x_{t-1};\mu_{\theta}\left( x_t,t \right),\Sigma_{\theta} \left( x_t,t \right) \right)$. 
In diffusion model, such a process to convert a data sample to a random noise sample (\textit{e.g.}, a sample from the standard Gaussian distribution) is referred to as the \textit{forward process} or the \textit{forward diffusion process}. If a neural network model can be used to estimate the inverse of the forward process (which is referred to as the \textit{reverse process} or the \textit{backward diffusion process}), then it can be used to generate data from noise. Formally, starting from the data sample $x_0$, a forward process of length $T$ can be constructed as follows.
\begin{equation*}
    q\left( x_{1:T}|x_0 \right):=\prod_{t=1}^{T} q\left( x_{t}|x_{t-1} \right), \quad q\left( x_{t}|x_{t-1} \right):=\mathcal{N}\left( x_{t-1};\sqrt{1-\beta_t}x_{t-1},\beta_{t}\textbf{I} \right)
\end{equation*}
where $\beta_1, ..., \beta_T$ are a sequence of scaling factors to scale the variance of noise added to each step of the forward process. 
In order to generate an authentic data sample $x_0$, the backward diffusion process implemented by the DDPM model needs to be trained to maximize the probability distribution $p_{\theta} \left( x_{0} \right)$, or equivalently to minimize the negative log-likelihood $-\log p_{\theta} \left( x_{0} \right)$. 
Ho \textit{et al.}\cite{ho2020denoising} showed that the negative log-likelihood term can be upper-bounded by the variational lower bound\cite{kingma2013auto}, from which the following model training objective can be derived by reparameterizing the Gaussian probability density function and ignoring the weighting terms containing $\beta_{t}$'s and $\Sigma_{\theta}$.
\begin{equation} \label{eq:Lt}
    L_{t}^{\text{simple}}=\mathbb{E}_{t\sim \left[1,T\right],x_0,\epsilon}\left[ {\|\epsilon_t-\epsilon_{\theta} \left( \sqrt{\Bar{\alpha}_t}x_0+\sqrt{1-\Bar{\alpha}_t}\epsilon_t,t \right)\|}^2 \right]
\end{equation}
where $\alpha_t:=1-\beta_t$, $\Bar{\alpha}_t:=\prod_{i=1}^{t}\alpha_i$, $\epsilon_t\sim\mathcal{N}\left(\textbf{0},\textbf{I}\right)$ is the standard Gaussian noise sampled at time $t$, and $\epsilon_{\theta}\left( \cdot \right)$ denotes the prediction of the DDPM neural network model given $\sqrt{\Bar{\alpha}_t}x_0+\sqrt{1-\Bar{\alpha}_t}\epsilon_t$ and $t$ as inputs. An illustration of the model training procedure is shown in the upper subplot of Figure \ref{fig:algorithm illustration}.
\subsection{Guided Data Synthesis with DDPM}


The original DDPM sampling algorithm\cite{ho2020denoising} generates data samples via a Markov chain starting from $x_{T} \sim\mathcal{N}\left(\textbf{0},\textbf{I}  \right)$. However, the randomness of $x_{T}$ makes it difficult to control the data generation process and therefore does not satisfy the need for high-fidelity CFD data reconstruction from a low-fidelity reference. To address this issue, a guided data generation procedure is needed where the low-fidelity CFD data is used as the condition to generate the high-fidelity one.
As suggested by Meng \textit{et al.}\cite{meng2021sdedit}, the Markovian property of the backward diffusion process means that the process to generate $x_0$ does not have to start from $x_T$ but can start from any time-step $t\in\{ 1,...,T \}$ provided that $x_t$ is available. This property allows users to select an intermediate sample at a particular time-step of the backward diffusion process, and send it to the remaining part of the Markov chain to obtain $x_0$. By controlling the signal component of the intermediate sample, the user can control the content of $x_0$. More specifically, let $x_0^{\text{forward}}$ denote the initial data sample at the beginning of the forward diffusion process, then the intermediate sample $x_t$ at an arbitrary time-step $t$ can always be calculated as follows.
\begin{equation}\label{eq:xt}
    x_t=\sqrt{\Bar{\alpha}_t}x_0^{\text{forward}}+\sqrt{1-\Bar{\alpha}_t}\epsilon_t
\end{equation}
Equation \ref{eq:xt} indicates that $x_t$ is a weighted combination of a signal component $x_0^{\text{forward}}$ and a random noise component $\epsilon_t$. Hence, for a guided data synthesis, one can assign a guidance data point $x^{\left(g\right)}$ to the signal component $x_0^{\text{forward}}$. The resulting Markov chain to generate a data sample $x_0$ conditioned on a reference $x^{\left(g\right)}$ is shown as follows.
\begin{equation}\label{eq:conditional-xt}
    p_{\theta}\left( x_{0:t} \right):=p\left( x_t \right) \prod_{i=1}^{t} p_{\theta}\left( x_{i-1}|x_{i} \right)\quad\text{where}\quad x_t:=\sqrt{\Bar{\alpha}_t}x^{\left(g\right)}+\sqrt{1-\Bar{\alpha}_t}\epsilon_t
\end{equation}
Note that the intermediate denoising result $x_t$ obtained from Eq. \ref{eq:xt} and Eq. \ref{eq:conditional-xt} are different  since $x_{0}^{\text{forward}}\neq x_{0}^{\left(g\right)}$. 
During model training, DDPM learns to generate a data sample $x_{0}$ using a Markov process starting from the $x_t$ in Eq. \ref{eq:xt}, such that $x_{0}$ approximates $x_{0}^{\text{forward}}$ (In the context of CFD data reconstruction, $x_{0}^{\text{forward}}$ is a high-fidelity CFD data sample from a training dataset, and $x_{0}$ is a reconstruction of $x_{0}^{\text{forward}}$ by DDPM). 
Nevertheless, when applying a trained DDPM model for conditional sampling, $x_{0}$ is generated using a Markov process starting from the $x_t$ in Eq. \ref{eq:conditional-xt}, given a guidance data point $x^{\left(g\right)}$. On one hand, we choose to use $x^{\left(g\right)}$ in data generation despite its difference from $x_{0}^{\text{forward}}$, because $x^{\left(g\right)}$ contains the conditioning information (\textit{e.g.}, a low-fidelity CFD data sample) which controls the backward diffusion process; on the other hand, by mixing the signal component ($x_{0}^{\text{forward}}$ or $x_{0}^{\left(g\right)}$) with random noise component $\epsilon_{t}$, we try to bring $\sqrt{\Bar{\alpha}_t}x^{\left(g\right)}+\sqrt{1-\Bar{\alpha}_t}\epsilon_t$ closer to $\sqrt{\Bar{\alpha}_t}x^{\text{forward}}+\sqrt{1-\Bar{\alpha}_t}\epsilon_t$ in the statistical sense (\textit{e.g.}, the distributions of the two quantities are both closer towards Gaussian), such that the disturbance to the quality of conditional sampling due to the discrepancy between $x_{0}^{\text{forward}}$ and $x^{\left(g\right)}$ will be mitigated.

\subsubsection{Improved Procedures for DDPM-based Conditional Sampling}
\begin{figure}
    \centering
    \includegraphics[width=0.85\textwidth]{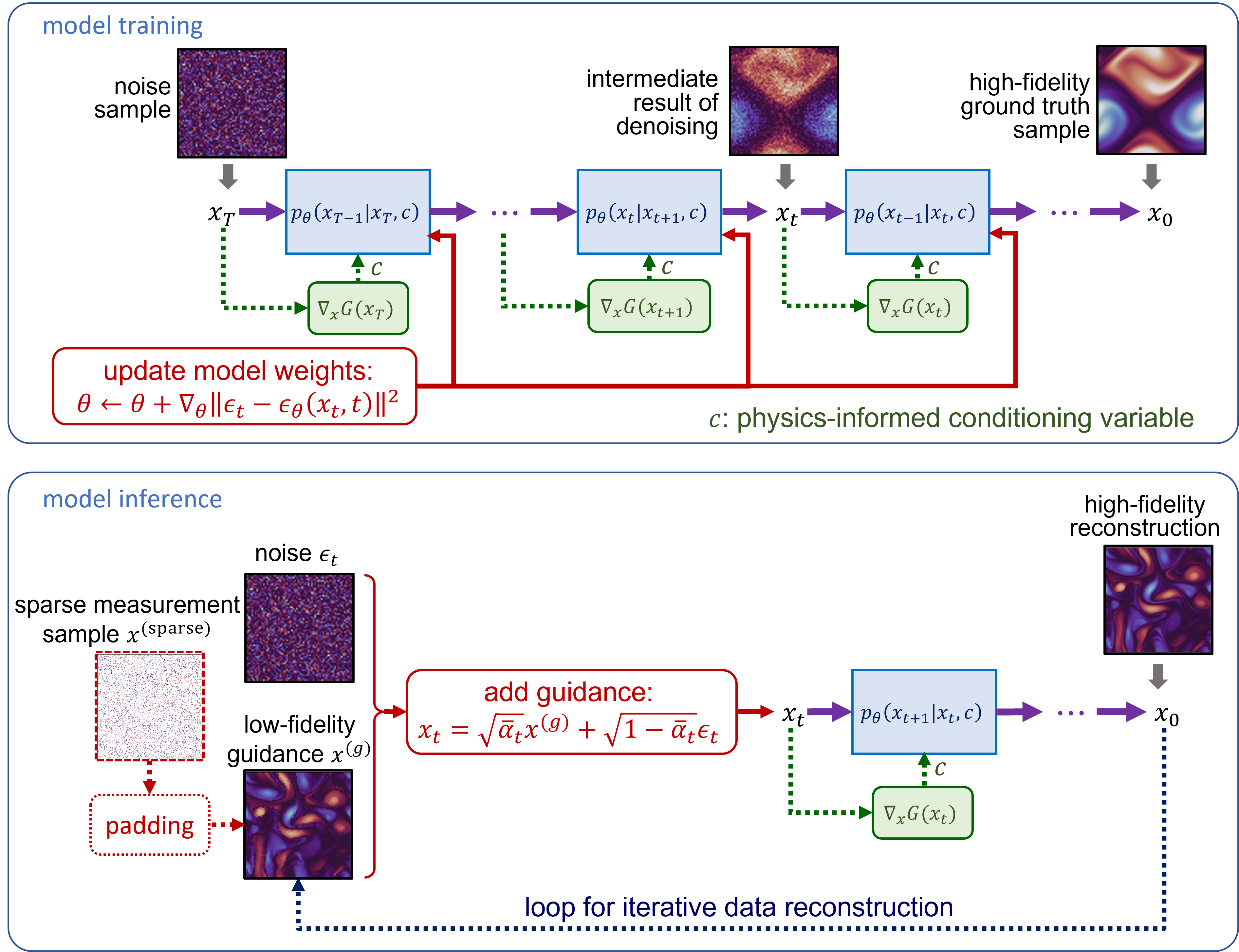}
    \caption{The procedures of model training (upper subplot) using Algorithm \ref{alg:ddpm-training-conditional} and model inference (lower subplot) using Algorithms \ref{alg:ddpm-conditional-sampling-conditional}. Each blue box represents a step in the backward diffusion process computed by a neural network model with parameter $\theta$. Blocks and arrows with dotted lines (\textit{e.g.}, sparsely measured samples and physics-informed conditioning) represents optional steps in the framework.}
    \label{fig:algorithm illustration}
\end{figure}

A related problem to the reconstruction of high-fidelity CFD data from low-fidelity input is the problem of reconstructing a high-fidelity data sample from an incomplete and sparsely measured data sample, as shown in the lower subplot of Figure \ref{fig:algorithm illustration}. 
Unmeasured components in a data sample are quite challenging to data reconstruction since they reduce the amount of information to use. One method to address this challenge is to add an iteration loop to the data reconstruction procedure. 
Given a sparsely measured data sample $x^{\left(\text{sparse}\right)}$, we first apply nearest-neighbor padding to fill the unmeasured data space, and then send the padded data sample to an iteration of DDPM-based conditional sampling procedure, where the reconstructed sample from the previous iteration is used as a low-fidelity reference to guide the next iteration of conditional sampling. 
This iteration can be repeated until the reconstruction quality has improved sufficiently. In general, the necessary number of iterations increases as the difference between $x_{0}^{\text{forward}}$ and $x^{\left(g\right)}$ becomes larger. 

The iterative sampling procedure above is one modification to the basic diffusion-based method made to address the special case of reconstructing high-fidelity CFD data from sparsely measured data samples. Additionally, we have considered a second special case where the analytical form of the Partial Differential Equation (PDE) determining the CFD data is known. To improve the backward diffusion process with the knowledge of the PDE, we propose another modification to the basic data sampling procedure using DDPM model.
Let the following equation be the known PDE operator that determines a fluid flow simulation from which the ground truth CFD data is collected,
\begin{equation}\label{eq:pde}
    G\left( u, \frac{\partial u}{\partial \xi_i}, \cdots, \frac{\partial^2 u}{\partial \xi_i \xi_j}, \cdots;\Theta \right)=0,\quad \xi = (\xi_1, \xi_2, \cdots) \in \Omega,
\end{equation}
where $G$ denotes the differential operator for the corresponding PDE, $u\left(\xi \right)$ is the solution to the PDE, $\Theta$ denotes the parameters in the PDE (e.g. viscosity, constant forcing), and $\Omega$ denotes the  computational domain on which the PDE is defined. 
If we substitute $u=x_t$ (where $x_t$ the intermediate step of the backward diffusion process) into the left-hand-side of Eq. \ref{eq:pde}, due to model prediction error and truncation error on the discretization grid, we will in general obtain a non-zero quantity on the right-hand-side of Eq. \ref{eq:pde}, \textit{e.g.}, $G|_{u=x_t}=r_t,\,\,r_t\not=0$. The non-zero term $r_t$ is usually referred to as the residual of the PDE and can be used to evaluate the accuracy of a numerical solution to the PDE. For conditional data generation, Ho \textit{et al.} \cite{ho2022classifier} proposed a classifier-free diffusion model that generates data samples from a conditional distribution $p_{\theta}\left( x_{t}|c \right)$ where $c$ is the class label of the data sample. Inspired by this work, we propose to use the gradient of the PDE residual as the conditioning information, which yields the following data sampling process. 
\begin{equation}\label{eq:conditional-xt-c}
    p_{\theta}\left( x_{0:t} \right):=p\left( x_t \right) \prod_{i=1}^{t} p_{\theta}\left( x_{i-1}|x_{i},c \right)\quad\text{where}\quad x_t=\sqrt{\Bar{\alpha}_t}x^{\left(g\right)}+\sqrt{1-\Bar{\alpha}_t}\epsilon_t,\quad c=\frac{\partial r_t}{\partial x}
\end{equation}
We refer to the conditioning information $c$ in Eq. \ref{eq:conditional-xt-c} as the physics-informed guidance, as it is obtained from the PDE which governs the physical process of the fluid flow. An overview of the model training and inference procedures for the physics-informed CFD data reconstruction is shown in Figure \ref{fig:algorithm illustration}.
Two methods are proposed in this work to implement the physics-informed data generation process shown in Eq. \ref{eq:conditional-xt-c}: 1. Data sampling with learned encoding of physics-informed guidance. 2. Data sampling with direct gradient descent of physics-informed guidance. In the first method, we modified the architecture of the original DDPM model by adding an extra module which encodes the PDE residual gradient $c=\partial r_t/\partial x_t$, so that the model can be used to sample $x_{i-1},\: i\in\{1,...,t\}$ from $p_{\theta}\left( x_{i-1}|x_{i},c \right)$ rather than $p_{\theta}\left( x_{i-1}|x_{i}\right)$. 
The modified model is trained using the following Algorithm \ref{alg:ddpm-training-conditional}.
\algdef{SE}[SUBALG]{Indent}{EndIndent}{}{\algorithmicend\ }%
\algtext*{Indent}
\algtext*{EndIndent}
\begin{center}
    \begin{minipage}{.75\linewidth}
      \begin{algorithm}[H]
        \caption{Physics-informed DDPM Model Training}\label{alg:ddpm-training-conditional}
        \begin{algorithmic}[1]
            \Require $p_{u}$ (probability of unconditional training), $G=0$ (PDE that determines the CFD data)
            \Repeat
            \State $x_0 \sim q\left( x_0 \right)$
            \State $t \sim \text{Uniform}\left( \{ 1,...,T \} \right)$
            \State $\epsilon_t \sim\mathcal{N}\left(\textbf{0},\textbf{I}\right)$
            \State $r_t=G|_{x=x_t}$ \Comment{Compute the residual of the PDE.}
            \State $c= 
                \begin{cases}
                    {\partial r_t}/{\partial x_t},& \text{with probability } 1-p_u\\
                    \emptyset,              & \text{with probability } p_u
                \end{cases}$ \Comment{Randomly discard conditioning information to train unconditionally.}
            \State Take a step of gradient descent on $\theta$ with the following gradient.
            \Indent
                $\nabla_{\theta} {\|\epsilon_t-\epsilon_{\theta} \left( \sqrt{\Bar{\alpha}_t}x_0+\sqrt{1-\Bar{\alpha}_t}\epsilon_t,t,c \right)\|}^2$
            \EndIndent
            \Until{converged}
        \end{algorithmic}
      \end{algorithm}
    \end{minipage}
\end{center}
Once the model is trained with Algorithm \ref{alg:ddpm-training-conditional}, it can be used in the data generation procedure specified by the following Algorithm \ref{alg:ddpm-conditional-sampling-conditional}.

\begin{center}
    \begin{minipage}{.9\linewidth}
      \begin{algorithm}[H]
        \caption{Physics-informed Conditional Sampling with DDPM }\label{alg:ddpm-conditional-sampling-conditional}
        \begin{algorithmic}[1]
            \Require $x^{\left(g\right)}$ (guide), $t$ (time-step location of $x^{\left(g\right)}$ in the backward diffusion process, $t<T$),
            $\tau=\{\tau_{0},\tau_{1},...,\tau_{S}\}$ (an increasing subsequence of $\left[0,1,...,t\right]$ where $\tau_{0}=0$ and $\tau_{S}=t$),
            $\epsilon_{\theta}$ (a pretrained DDPM model), $G=0$ (PDE that determines the CFD data), $w$ (guidance strength)
            \For {$k=1,2,...,K$}
                \State $\epsilon_t \sim \mathcal{N}\left(\textbf{0},\textbf{I}\right)$
                \State $x_t=\sqrt{\Bar{\alpha}_t}x^{\left(g\right)}+\sqrt{1-\Bar{\alpha}_t}\epsilon_t$
                
                \For {$i=S,S-1,...,1$}
                    \State $z \sim \mathcal{N}\left(\textbf{0},\textbf{I}\right)$ if $i>1$ else $z=0$
                    \State $r_{\tau_i}=G|_{u=x_{\tau_i}}$ \Comment{Compute the residual of the PDE.}
                    \State $c={\partial r_{\tau_i}}/{\partial x_{\tau_i}}$
                    \State $\Tilde{\epsilon}_{\theta}=\epsilon_{\theta}\left( x_{\tau_i},{\tau_i},c \right)+w\left[\epsilon_{\theta}\left( x_{\tau_i},{\tau_i},c \right)-\epsilon_{\theta}\left( x_{\tau_i},{\tau_i},\emptyset \right) \right]$
                    \State $x_{\tau_{i-1}}=\frac{\sqrt{\Bar{\alpha}_{\tau_{i-1}}}}{\sqrt{\Bar{\alpha}_{\tau_i}}}
                    \left( x_{\tau_i}-\sqrt{1-\Bar{\alpha}_{\tau_i}}\cdot\Tilde{\epsilon}_{\theta} \right)+
                    \sqrt{1-\Bar{\alpha}_{\tau_{i-1}}-\sigma_{{\tau_i}}^{2}}\cdot\Tilde{\epsilon}_{\theta}+\sigma_{{\tau_i}}^{2}\epsilon_{\tau_i}$
                \EndFor
                \State $x^{\left(g\right)} = x_{0}$ \Comment{$x_{\tau_0}=x_{0}$ since $\tau_0:=0$.}
            \EndFor
            \State \Return $x_0$
        \end{algorithmic}
      \end{algorithm}
    \end{minipage}
\end{center}
\medskip
\medskip
\medskip
In the context of high-fidelity CFD data reconstruction, $x^{\left(g\right)}$ represents a sample of low-fidelity CFD data, $x_0$ represents the corresponding sample of high-fidelity CFD data, the DDPM model $\epsilon_{\theta}$ is pretrained on a high-fidelity CFD dataset, and $t$ is a main hyper-parameter for conditional data sampling using Algorithm \ref{alg:ddpm-conditional-sampling-conditional}. In practice, $t$ is selected from the interval $\left[0, \frac{T}{2}\right]$ for a more accurate and less noisy data reconstruction. 
Instead of following the data sampling procedure from the original DDPM model, we designed Algorithm \ref{alg:ddpm-conditional-sampling-conditional} based on the accelerated sampling procedure proposed by Song \textit{et al.}\cite{song2020denoising}, named the Denoising Diffusion Implicit Model (DDIM). We also adopted the same design choice $\sigma_{\tau_i} = 0 \;\forall\;i$ as suggested in the original DDIM model.
The intuition behind Line 8 in Algorithm \ref{alg:ddpm-conditional-sampling-conditional} is briefly discussed as follows. Ho \textit{et al.} pointed out in their work \cite{ho2020denoising} that the data sampling process presented in 
the original DDPM resembles the Langevin dynamics with $\epsilon_{\theta}$ as a learned gradient of the data density, referred to as the score function\cite{score-based-gen}. The resemblance between $\epsilon_{\theta}$ and the score function enables the following approximation.
\begin{equation}\label{eq:score-approximation}
    \epsilon_{\theta}\left( x_i,i,\emptyset \right)\approx -\sigma_{i} \nabla_{x_i} \log p_{\theta} \left(x_{i}\right),\quad \epsilon_{\theta}\left( x_i,i,c \right)\approx -\sigma_{i} \nabla_{x_i} \log p_{\theta} \left(x_{i},c\right)
\end{equation}
Eq. \ref{eq:score-approximation} indicates that the neural network model $\epsilon_{\theta}$ in the DDPM and DDIM models can be considered as an estimator of the score function. Substitute Eq. \ref{eq:score-approximation} into Line 8 of Algorithm \ref{alg:ddpm-conditional-sampling-conditional}, we have
\begin{equation}\label{eq:epsilon-expand}
\begin{split}
    \Tilde{\epsilon}_{\theta} &= \epsilon_{\theta}\left( x_i,i,c \right)+w\left[\epsilon_{\theta}\left( x_i,i,c \right)-\epsilon_{\theta}\left( x_i,i,\emptyset \right) \right]\\
    & \approx -\sigma_{i} \nabla_{x_i} \log p_{\theta} \left(x_{i},c\right) - w\left[ \sigma_{i} \nabla_{x_i} \log p_{\theta} \left(x_{i},c\right) -\sigma_{i} \nabla_{x_i} \log p_{\theta} \left(x_{i}\right) \right] \\
    & = -\sigma_{i} \nabla_{x_i} \log p_{\theta} \left(x_{i},c\right) -w\sigma_{i} \nabla_{x_i} \log p_{\theta} \left(c|x_{i}\right)
\end{split}
\end{equation}
Equation \ref{eq:epsilon-expand} shows that Algorithm \ref{alg:ddpm-conditional-sampling-conditional} is designed to sample $x_{i-1}$ using a weighted combination of the data prediction $p_{\theta} \left(x_{i},c\right)$ and the PDE residual gradient prediction $p_{\theta} \left(c|x_{i}\right)$, where the weight is determined by the guidance strength $w$.
In our second method to implement the physics-informed data generation process, data sampling with direct gradient descent of physics-informed guidance, we do not modify the original DDPM model architecture or training procedure to accommodate PDE residual gradient $c$ in model input. Instead, we directly incorporate gradient descent in the conditional sampling process of DDIM model. The intuition for this method is as follows. Consider the PDE in Eq. \ref{eq:pde} which models the fluid flow. Any ground truth CFD data sample should satisfy this PDE and produce a zero residual. Therefore, a valid goal of high-fidelity CFD data reconstruction is to optimize the DDPM model preditction $x_0$, such that the corresponding residual $r_{0}:=G|_{u=x_0}$ is minimized. A gradient-descent method to achieve this goal is to search for the high-fidelity CFD data using the following update rule, $x_{i-1}=x_{i}-\lambda\partial r_{i}/\partial x_{i}$, where $\lambda$ represents the step size of gradient descent. For an improved performance of high-fidelity CFD data reconstruction, we combined the original goal of DDPM-based data sampling (which is to minimize the KL-divergence between the forward and the backward diffusion processes for an authentic data generation) with the goal of minimizing the residual of CFD data reconstruction. More specifically, to implement our data sampling method with direct gradient descent of physics-informed guidance, we modified Line 8 of Algorithm \ref{alg:ddpm-conditional-sampling-conditional} as follows.
\begin{equation}
    x_{\tau_{i-1}}=\frac{\sqrt{\Bar{\alpha}_{\tau_{i-1}}}}{\sqrt{\Bar{\alpha}_{\tau_i}}}
                    \left( x_{\tau_i}-\sqrt{1-\Bar{\alpha}_{\tau_i}}\cdot\Tilde{\epsilon}_{\theta} \right)+
                    \sqrt{1-\Bar{\alpha}_{\tau_{i-1}}-\sigma_{{\tau_i}}^{2}}\cdot\Tilde{\epsilon}_{\theta}-\lambda c+\sigma_{{\tau_i}}^{2}\epsilon_{\tau_i}
\end{equation}


\section{Experiments}

\subsection{Implementation}

The dataset we considered is the 2-dimensional Kolmogorov flow \cite{chandler_2013_kolmogorov}, governed by the Navier-Stokes equation for incompressible flow. The vorticity form of it reads as,
\begin{equation}
    \begin{aligned}
\frac{\omega (\mathbf{x},t)}{\partial t} + \mathbf{u}(\mathbf{x},t) \cdot \nabla \omega (\mathbf{x},t) &= \frac{1}{\textit{Re}} \nabla^2 \omega (\mathbf{x},t) + f(\mathbf{x}) , \quad \mathbf{x} \in (0,2\pi)^2,  t \in (0,T], \\
\nabla \cdot \mathbf{u}(\mathbf{x},t) &= 0, \quad  \mathbf{x} \in (0,2\pi)^2, t \in (0,T], \\
\omega (\mathbf{x},0) &= \omega_0(\mathbf{x}), \quad \mathbf{x} \in (0,2\pi)^2,
\end{aligned}
\label{kmflow-eq}
\end{equation}
where $\omega$ is the vorticity, $\mathbf{u}$ represents the velocity field, $\textit{Re}$ denotes the Reynolds number which is set to $1000$ in this problem, $f(\mathbf{x})$ is the forcing term, and $\mathbf{x}=[x_1, x_2]$. The boundary condition considered is periodic boundary condition. The forcing term for 2-d Kolmogorov flow studied here is defined as: $f(\mathbf{x}) = -4 \cos(4x_2) - 0.1 \omega(\mathbf{x}, t)$, which contains an additional drag force term $0.1 \omega(\mathbf{x}, t)$ similar to Kochkov \textit{et al.}\cite{Kochkov_2021}, in order to prevent energy accumulation at large scales\cite{2d-turbulence_guido2012}. We numerically solve Equation \eqref{kmflow-eq} using a pseudo-spectral solver implemented in PyTorch \cite{pytorch} from Li \textit{et al.} \cite{PINO_li_2021}, which samples initial condition $\omega_0(\mathbf{x})$ from a Gaussian random field $\mathcal{N}(0, 7^{3/2}(-\Delta+49I)^{-5/2})$. The discretization grid used is a $2048 \times 2048$ uniform grid. The data derived from the direct numerical simulation are considered the high-fidelity data we aim to reconstruct using the machine learning model, and are referred to as ground truth in the dataset.

We simulate 40 sequences in total, each with a temporal length of 10 seconds ($T=10$). We then downsample these data to a $256 \times 256$ grid spatially with a fixed time interval $\Delta t=1/32 s$, resulting in a dataset comprising 40 sequences each with 320 frames. Among them, we use the first 36 sequences as the training set and the rest 4 for testing. As our model operates on the same input and output grid, we use nearest (based on Euclidean norm) interpolation to process the input such that it has the same resolution as the target.

For the calculation of the PDE's residual, we use discrete Fourier transform to calculate spatial derivatives and finite difference to calculate time derivatives. Given that the proposed diffusion model operates on the vorticity of three consecutive frames $[\omega_{t-1}(\mathbf{x}), \omega_{t}(\mathbf{x}), \omega_{t+1}(\mathbf{x})]$, we can approximate $\partial_t \omega $ via $ \partial_t \omega \approx (\omega_{t+1}(\mathbf{x}) - \omega_{t-1}(\mathbf{x})) / (2\Delta t)$. For the convection term and diffusion term, we approximate them by calculating the Laplacian and gradient of vorticity in the Fourier space,  derive velocity using the stream function $\psi$: $\mathbf{u} = \nabla \times \psi, \psi=\nabla^{-2} \omega$, and then convert them back to the physical space via inverse Fourier transform.

The proposed denoising diffusion model is parameterized by a UNet \cite{UNet}, which has empirically been shown effective for estimating the score function\cite{ho2020denoising, score-based-gen} (or equivalently the noise in DDPM). UNet is a neural network architecture with hierarchical convolution blocks and multi-level skip connections, which allows it to better capture the dependency of different ranges (resembles multigrid method). In addition, we use self-attention\cite{attention} in the bottleneck layer (corresponds to the coarsest resolution) to further increase the receptive field. Based on the UNet architecture, we investigate three types of training strategies as below.
\begin{itemize}
    \item  Diffusion model with learned residual guidance. The residual information is directly provided to the diffusion model as input features, i.e. $\epsilon_\theta\left[x_t, \nabla_{x_t}G\left(x_t\right)\right]$, where $G(\cdot)$ is the corresponding differential operator of the Navier-Stokes equation. In this way, the final guidance is the learned combination between residual gradient and score function.
    \item Original diffusion model. In this case, the diffusion process will not take residual information into account.
    \item A UNet that learns a fixed mapping: $f: X \mapsto Y$ where $X$ is the low fidelity data domain and $Y$ is the high fidelity data domain. This is the setup which most deep-learning-based flow reconstruction methods have adopted \cite{pant2020deep, yousif2021high, fukami2019super, fukami2021machine}. This setup requires paired training data and the learned mapping is usually locked to a specific distribution in $X$.
\end{itemize}

We adopt a recursive refinement strategy for which we apply $K=3$ (see Algorithm \ref{alg:ddpm-conditional-sampling-conditional} for detailed definition of $K$) times of backward diffusion process recursively for $4\times$ upsampling and $5\%$ points' sparse reconstruction tasks, and we set $S=160, 114, 80$ for $k=1, 2, 3$ respectively. However, for input data that is farther away from the target data distribution, we observe that a single diffusion chain with larger injected noise (i.e. larger $t$ in line 3 of Algorithm \ref{alg:ddpm-conditional-sampling-conditional}) is often more beneficial. Therefore, for $8\times$ upsampling task we use $S=320$ and set $K=1$, and $S=400, K=1$ for $1.5625\%$ reconstruction task.
\section{Results}

We conduct experiments on the following tasks to investigate the capability of the diffusion model on reconstructing high fidelity flow field. The task in the first experiment is reconstructing high-resolution field from low-resolution field, where the low-resolution field is uniformly downsampled from the high-resolution one. The task in the second experiment is to reconstruct high-resolution field from randomly sampled collocation points (not necessarily equidistant). The second task aims to reconstruct dense field from sparse sensory observation data. For the first task, we test our reconstruction model on two levels of input resolution, $64 \times 64 \mapsto 256 \times 256$ ($4\times$ upsampling) and $32 \times 32 \mapsto 256 \times 256$ ($8\times$ upsampling). For the second task, we also test our reconstruction model on two different levels of sparsity, where we sampled $5\%$ and $1.5625\%$ (same amount of grid points as $32 \times 32$ grid). The collocation points is randomly sampled with uniform probability. Note that for all the experiments (except the ablation for different guidance methods), we use the same diffusion model (with learned guidance), which means we do not retrain the model for a specific task. To reduce the aliasing effect when applying direct mapping model to out-of-distribution data, we applied Gaussian smoothing kernel with $\sigma=5$ to data used in $32 \times 32 \mapsto 256 \times 256$ and $1.5625\% \mapsto 256 \times 256$ tasks.

\begin{figure}[H]
    \centering
    \includegraphics[width=\linewidth]{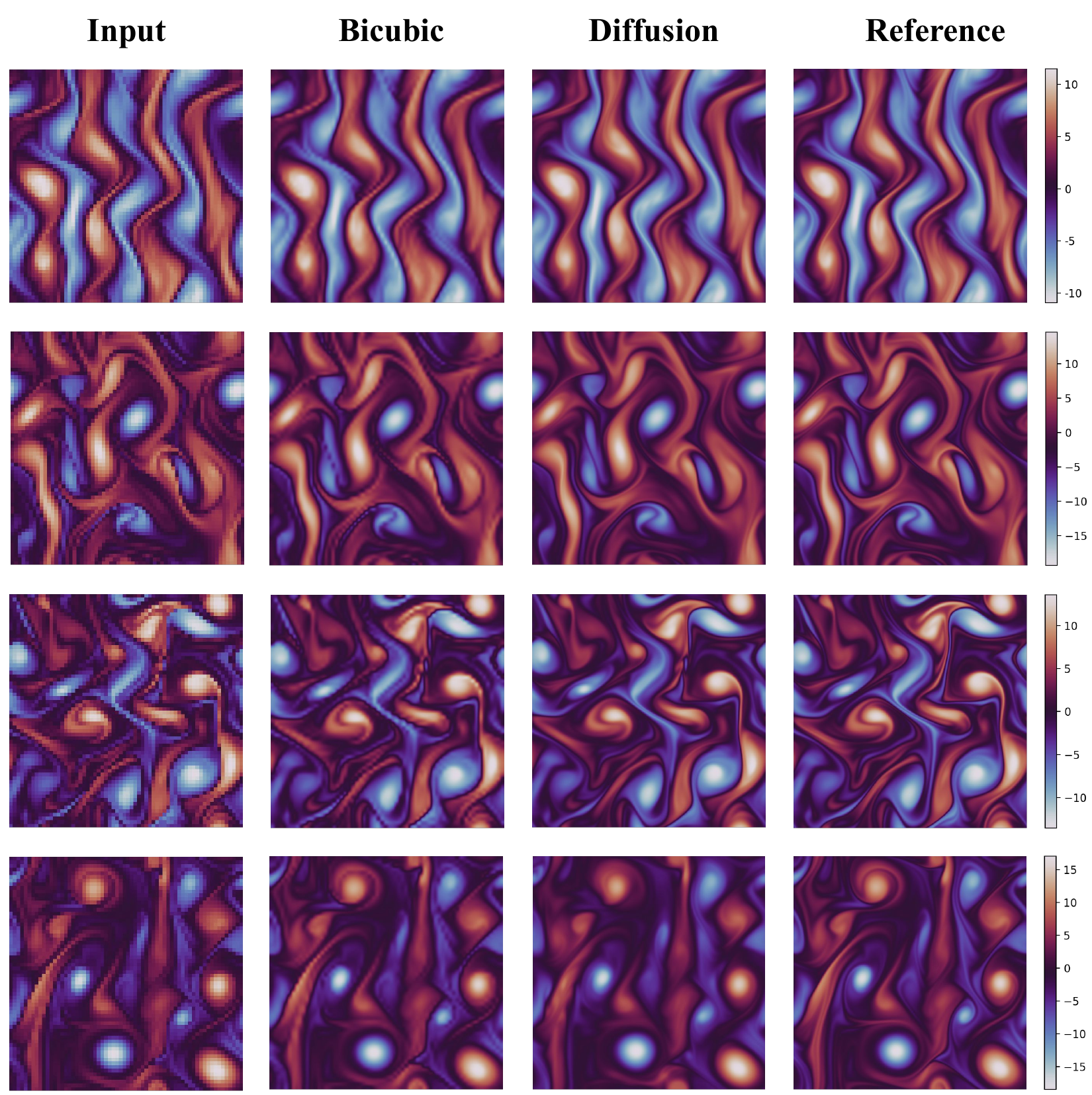}
    \captionof{figure}{Qualitative comparison of different upsampling methods on 4x upsampling task.}
    \label{vis-4x}
\end{figure}

\begin{figure}[H]
    \centering
    \includegraphics[width=\linewidth]{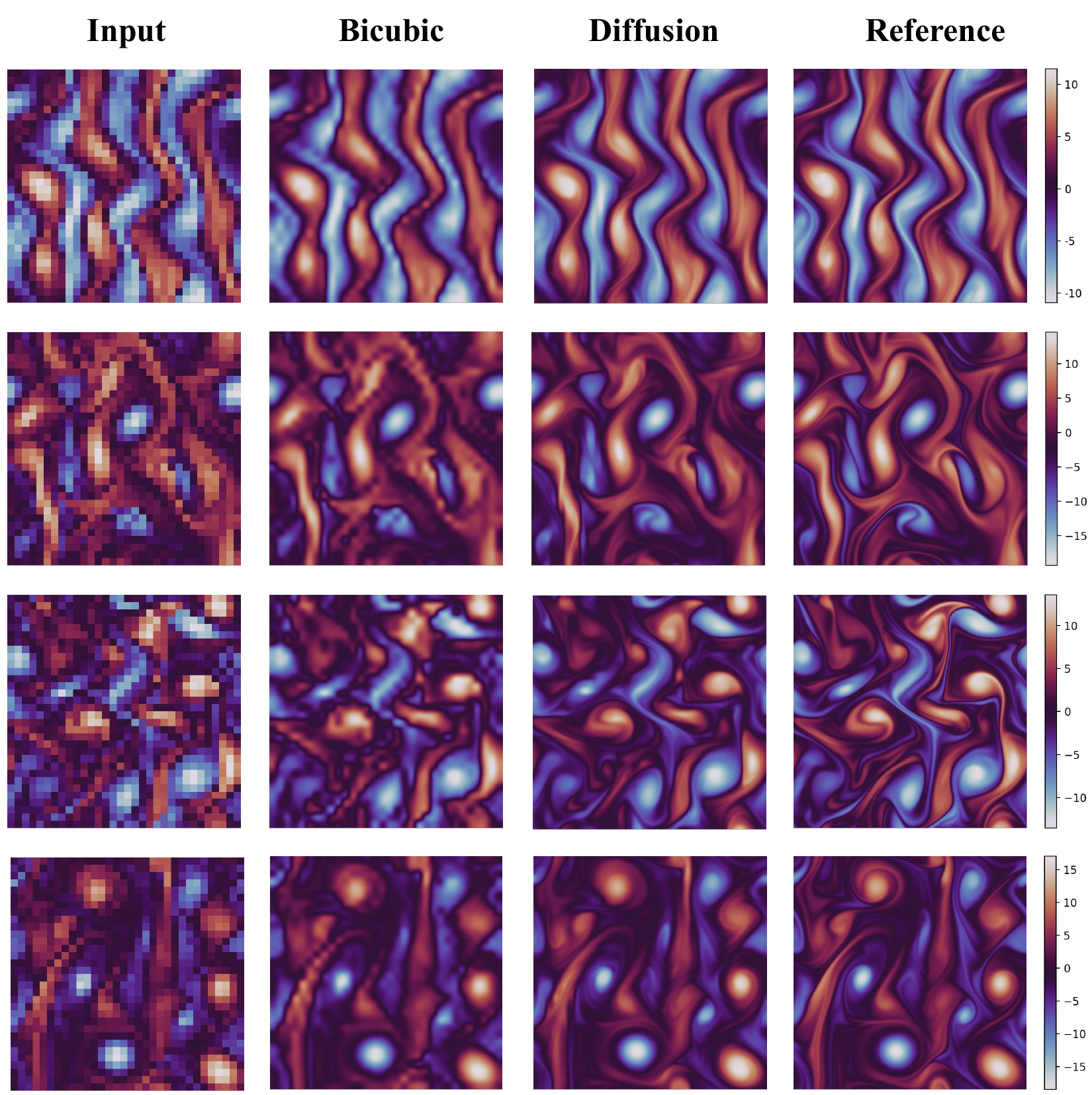}
    \captionof{figure}{Qualitative comparison of different upsampling methods on 8x upsampling task.}
    \label{vis-8x}
\end{figure}

The visualization of our model's reconstruction results are shown in the Figures \ref{vis-4x}, \ref{vis-8x}, \ref{vis-sparse}. We can observe that both bicubic interpolation and diffusion model generate satisfactory  results for the easier task of $4\times$ upsampling, but the diffusion model can recover more details for the more challenging scenarios (Figure \ref{vis-8x}, \ref{vis-sparse}). This qualitatively demonstrates the capability of using diffusion model to reconstruct high-resolution flow field given inputs with different distributions.

\begin{figure}[H]
    \centering
    \includegraphics[width=\linewidth]{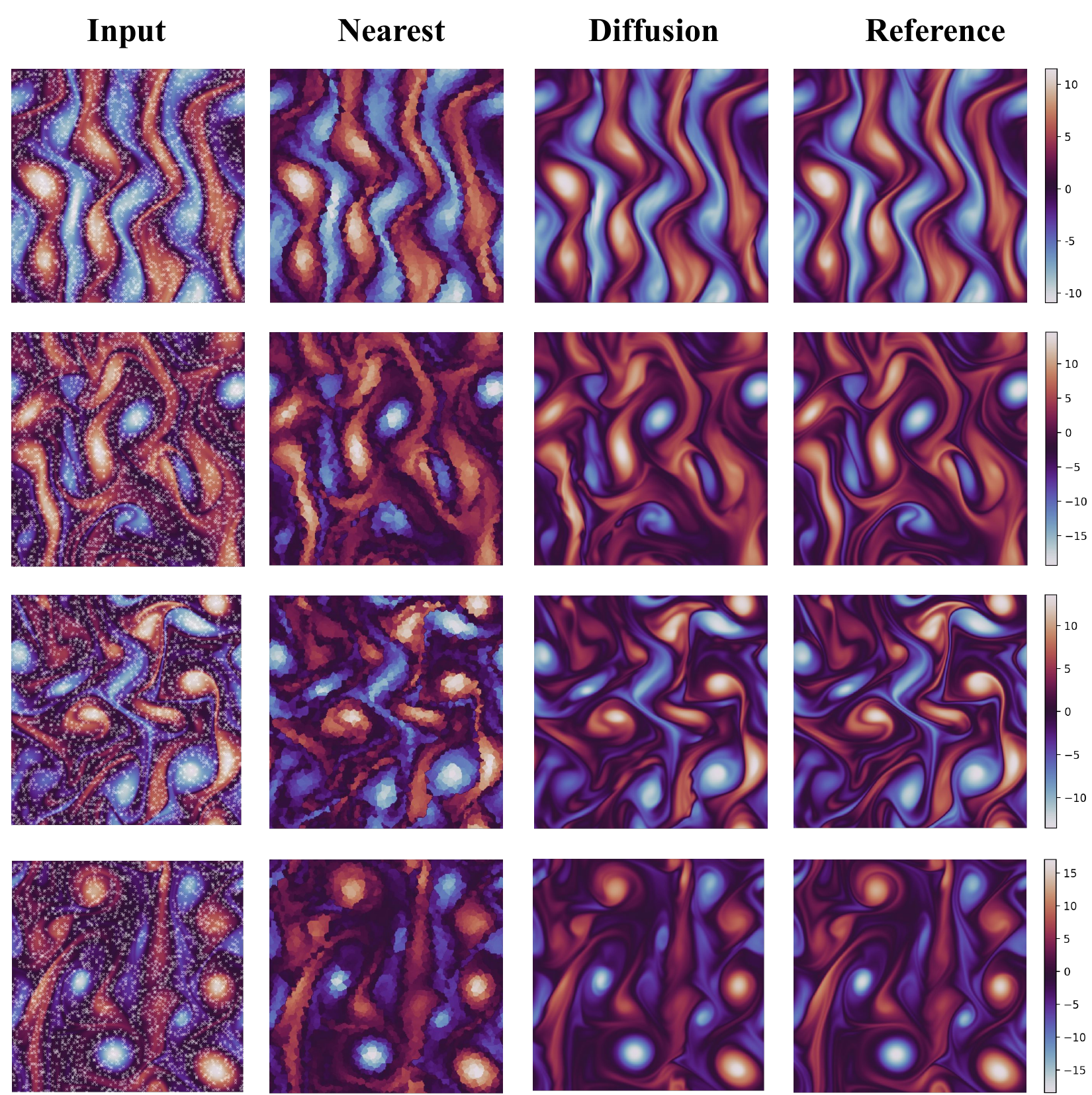}
    \captionof{figure}{Qualitative comparison of different upsampling methods on non-equidistant sparse reconstruction task using $5\%$ of grid points. White cross in the input indicates the collocation points (zoom in for clarity).}
    \label{vis-sparse}
\end{figure}
\begin{figure}[H]
    \centering
    \includegraphics[width=\linewidth]{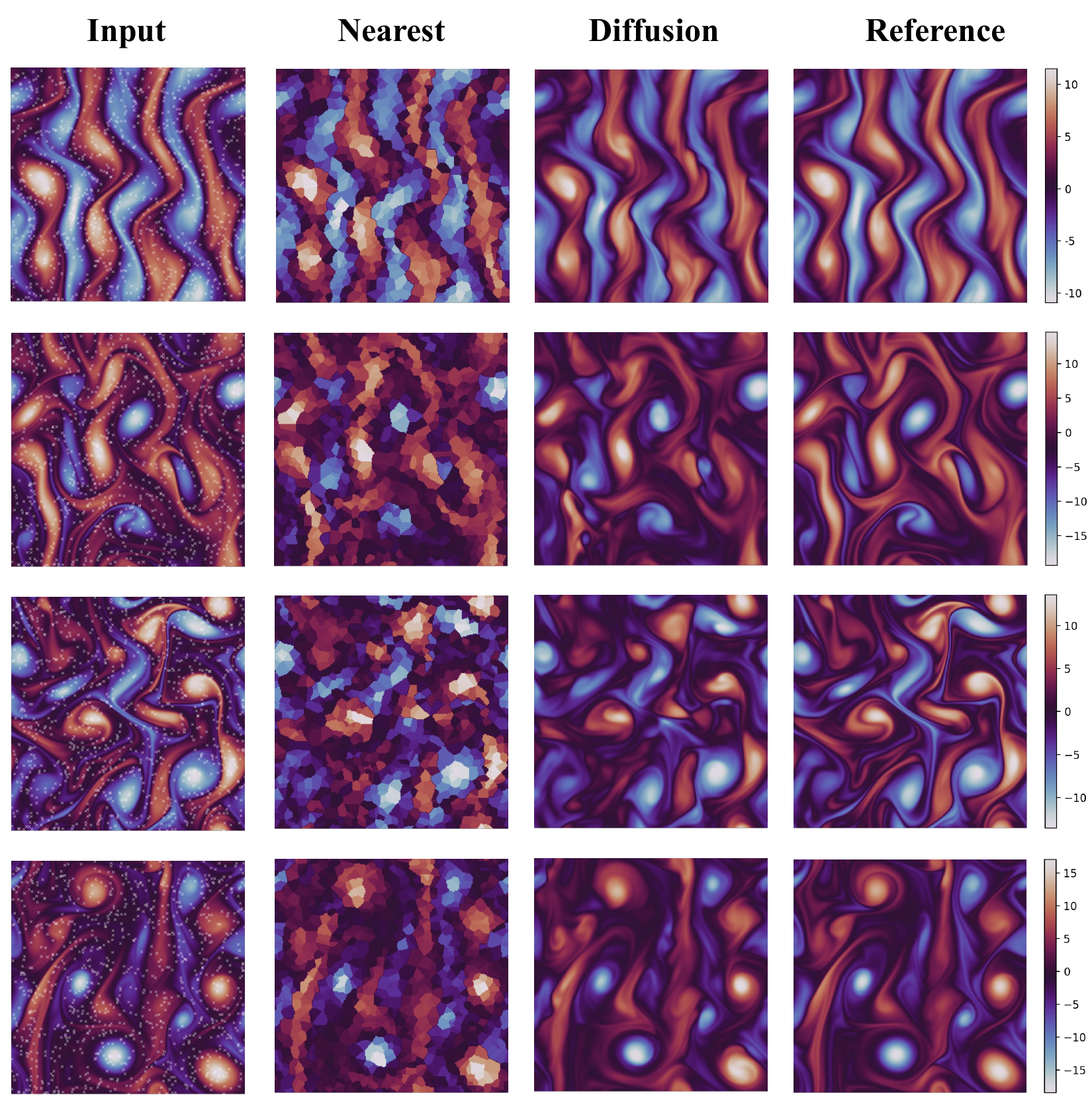}
    \captionof{figure}{Qualitative comparison of different upsampling methods on non-equidistant sparse reconstruction task using $1.5625\%$ of grid points. White cross in the input indicates the collocation points (zoom in for clarity).}
    \label{vis-sparse}
\end{figure}
\newpage

To quantitatively evaluate the reconstruction results, we use $L_{2}$ norm to measure the pointwise error between prediction and ground truth on each grid point (with $n$ representing the total number of grid points per sample):
\begin{equation}
    D_{L_{2}}(\hat{\omega}, \omega) = \sqrt{\frac{1}{n}\sum_{i=1}^n(\hat{\omega_i}-\omega_i)^2}.
\end{equation}
We also evaluate the normalized residual of each predicted result that comprise three consecutive frames. This measures if the results are aligned with the underlying governing equation.
\begin{equation}
    D_{\text{equation}}(\hat{\omega}, \omega) = \frac{1}{n}\frac{\sum_{i=1}^n{|G(\hat{\omega_i})|^2}}{||\omega||_2^2},
\end{equation}
where $G(\cdot)$ is the differential operator as described in Equation \eqref{kmflow-eq}, $||\omega||_2^2$ is the average squared $L_{2}$ norm of the vorticity.\\
In addition to the pointwise metric, we compare the statistics between the reconstructed results and the ground truth. More specifically, we inspect the energy spectrum and vorticity distribution to validate the alignment between prediction and ground truth distribution. 

The pointwise error of different models' reconstruction results are shown in the Table \ref{tab1-quant}. Both the diffusion model and the learned direct mapping model outperform bicubic interpolation by a margin, which indicates the strength of data-driven learning-based methods. In terms of $L_{2}$ loss, direct mapping model has similar performance as diffusion model, yet the margin enlarges when applying the direct mapping model to a new data distribution (mapping from observation on $5\%$ of the grid points to full resolution grid). Both the $L_{2}$ loss and the equation loss increase significantly for the direct mapping model, while the diffusion model shows more robustness by outperforming the direct mapping model. This signifies the difference between the two learning paradigms. Direct mapping model is sensitive with respect to the input distribution, while in diffusion model, the input is perturbed with noises and thus the perturbed input distribution still intersects with the distribution of which the diffusion model can effectively estimate the score function. As shown in Figure \ref{vis-dm-vs-diffusion}, the learned direct mapping model's prediction is much blurrier when applied to out-of-distribution data. Furthermore, with physics-informed guidance, the PDE residual of diffusion model's predicted results are lower than other models that do not take PDE information into account.

\begin{figure}[H]
    \begin{subfigure}{0.49\textwidth}
        \includegraphics[width=\linewidth]{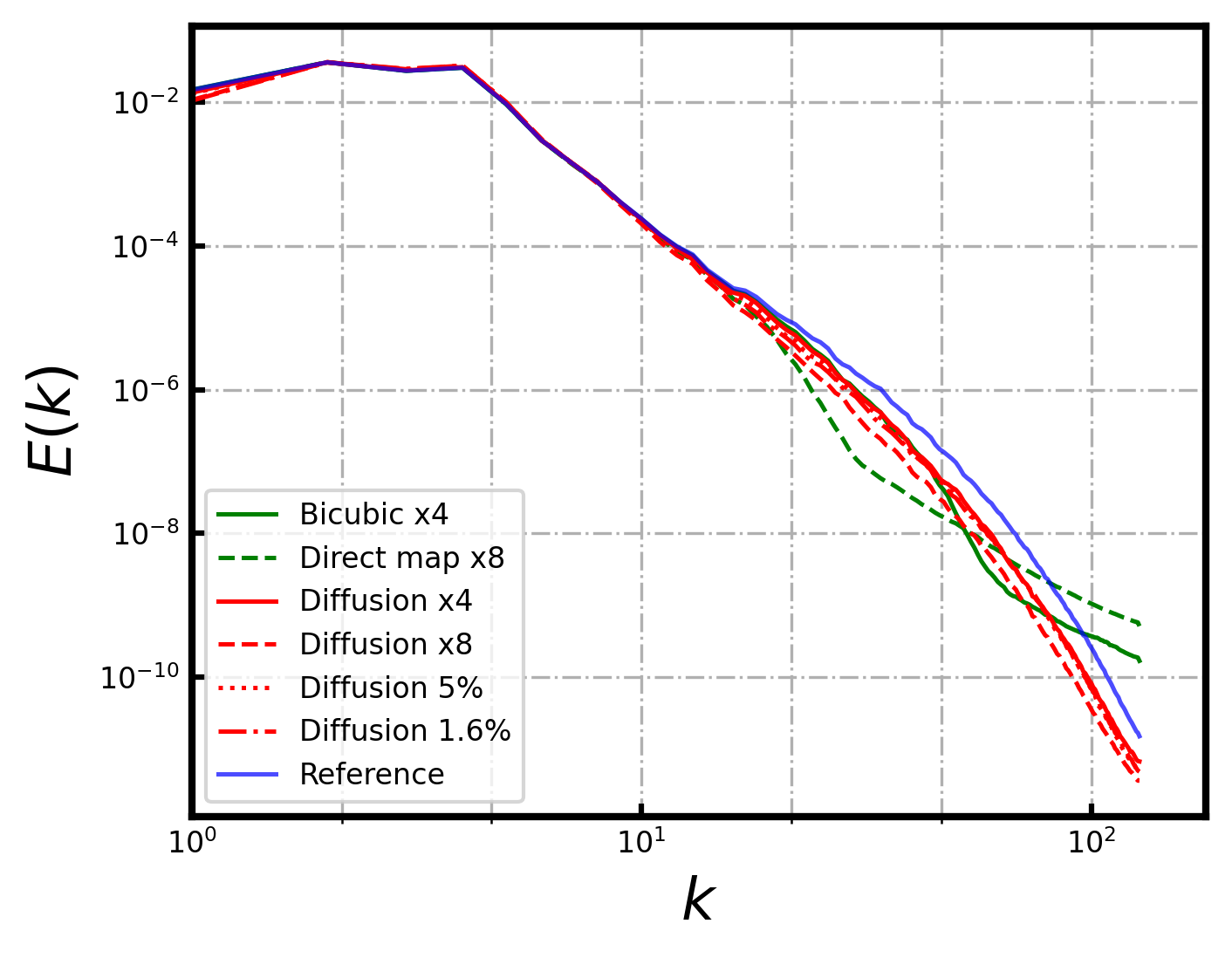}
        \captionof{figure}{Kinetic energy spectrum}
    \end{subfigure}
    \begin{subfigure}{0.5\textwidth}
        \includegraphics[width=\linewidth]{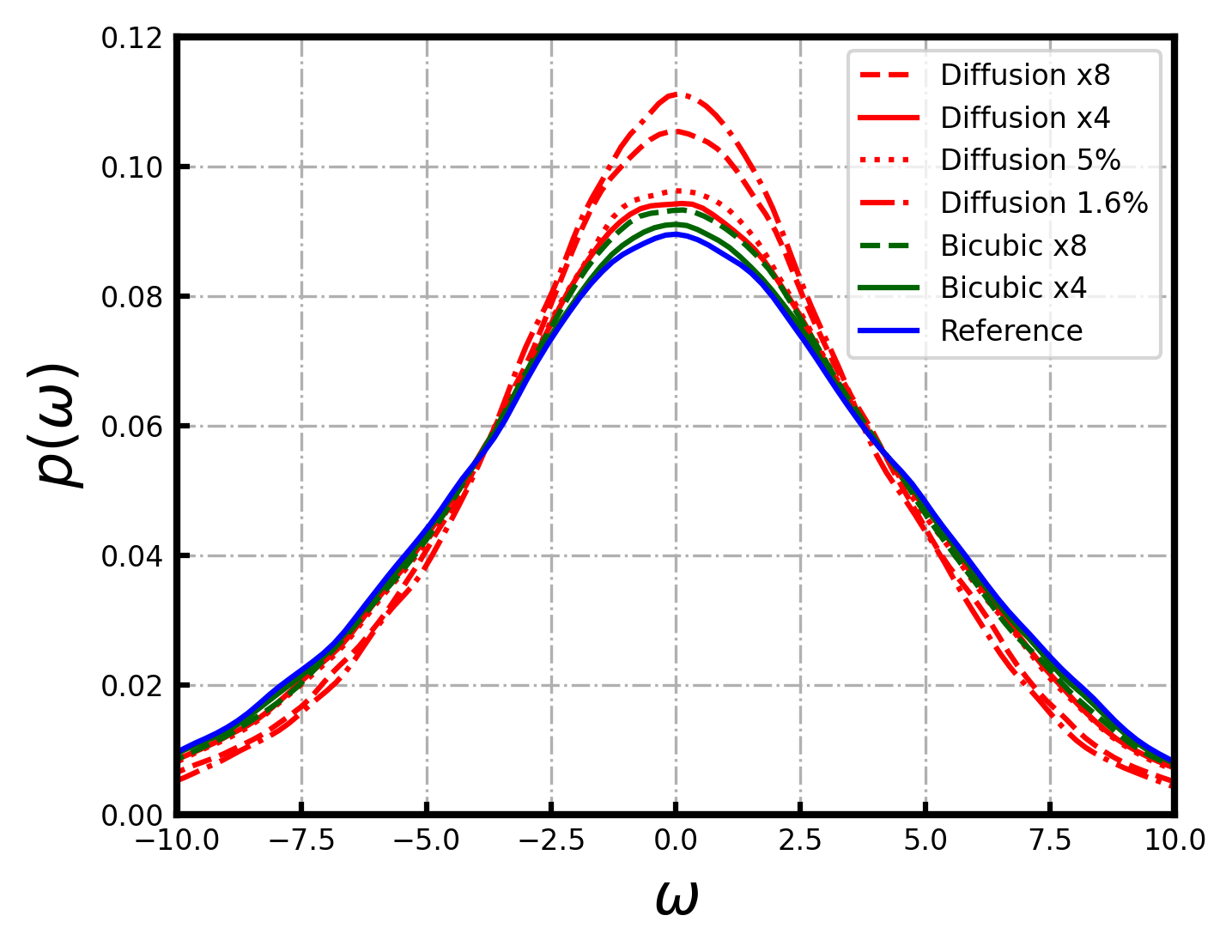}
        \captionof{figure}{Vorticity distribution \label{vis-vort-comp}}
    \end{subfigure}
    \caption{Statistics of different reconstruction methods and reference result.}
\end{figure}

Next we compare the voriticity distribution between the predicted data and the ground truth. Figure \ref{vis-vort-comp} shows that the results of the diffusion model have a sharper distribution than both the ground truth and bicubic interpolated data. This is likely caused by the noise estimation error of the diffusion model. Before the backward diffusion process starts, the input data is mixed with Gaussian noise, then the diffusion model gradually removes the noise and steps towards the target distribution. However, as the noise estimation from diffusion model always contains error to some degrees, the final result is not guaranteed to be perfectly noise-free and exhibits a slight distribution drift. This is more obvious for upsampling case with higher sparsity, where we inject the input data with larger noise and diffuse for more steps. For the energy spectrum, compared to bicubic interpolation, we observe that diffusion model can better capture the trend in high wavenumber regime and exhibits less truncation effect. As for the direct mapping model, it has reasonable accuracy on the data it was trained on. However, when extrapolated to the out-of-distribution data, its performance degrades and produces results that has much higher residual. As shown in Figure \ref{vis-dm-vs-diffusion}, its prediction becomes non-smooth and less physical coherent.

PDE's residual calculation is sensitive with respect to noise and aliasing (due to nearest padding). To accommodate for this, for task with high sparsity ($1.5625\%$), we apply Gaussian filter to smooth the field before inputting them to the model. From Table \ref{tab2-quant-filter} and Figure \ref{vis-dm-vs-diff-energy}, we observe that while smoothing can alleviate direct mapping model's degradation, it still struggles to predict physically coherent result compared to diffusion model. Furthermore, despite a smoothing process can help improve prediction's $L_2$ norm, a low-pass filter like Gaussian filter will lose energy in higher frequency regime (as shown in Figure \ref{vis-dm-vs-diff-energy}). 

\begin{table}[H]
\centering
\begin{tabular}{c|ccc|ccc} 
\toprule
\multirow{2}{*}{Task}                  & \multicolumn{3}{c|}{$L_{2}$ norm} & \multicolumn{3}{c}{Equation loss}  \\ 
\cmidrule{2-7}
& Diffusion & Direct map & Bicubic  & Diffusion & Direct map & Bicubic   \\ 
\midrule
$64 \times 64 \mapsto 256 \times 256 $ & 0.5622    & 0.6048     & 1.3355   & 0.2178    & 0.5438     & 10.6639   \\ 
\cmidrule{1-7}
$32 \times 32 \mapsto 256 \times 256 $ & 1.3035    & 1.3700     & 2.5179   & 0.2039    & 9.9291     & 20.0008   \\
$5 \% \mapsto 256 \times 256 $         & 0.9318    & 1.2426     & -        & 0.2815    & 20.2853    & -         \\
$1.6 \% \mapsto 256 \times 256 $ & 1.8213    & 1.9149   & -             & 0.2290    & 17.5249 & - \\
\bottomrule
\end{tabular}
\caption{Quantitative comparison of diffusion model and other interpolation/reconstruction methods. The "Direct map" model is trained on $64 \times 64 \mapsto 256 \times 256 $ data, data in other tasks are considered out-of-distribution data for it. \label{tab1-quant}
}
\end{table}

\begin{figure}[H]
    \begin{subfigure}{0.85\textwidth}
        \includegraphics[width=\linewidth]{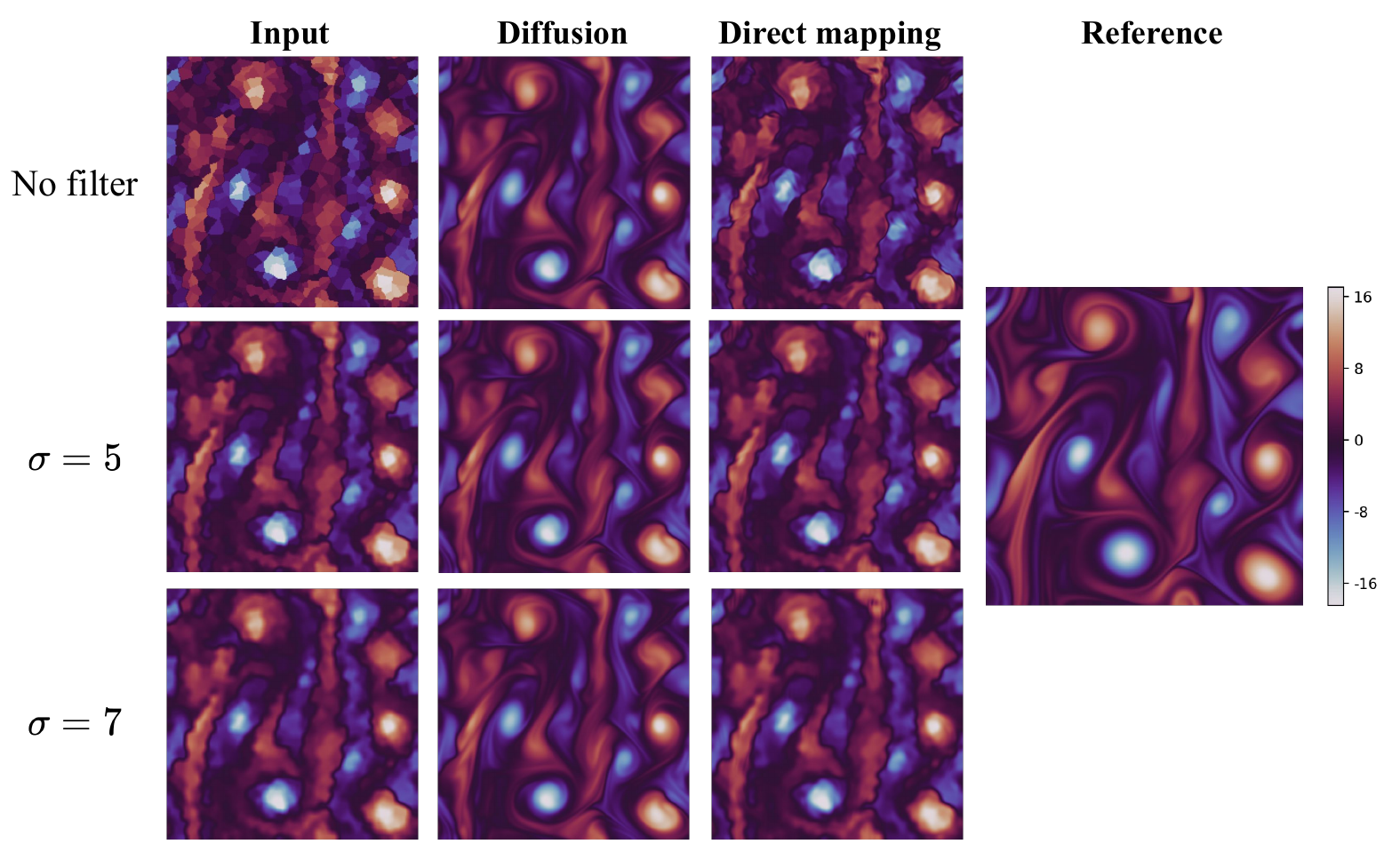}
        \captionof{figure}{Qualitative comparison of different models' prediction on the sparse reconstruction task with different filtered input. \label{vis-dm-vs-diffusion}}
    \end{subfigure}
\end{figure}%
\begin{figure}[H]\ContinuedFloat
    \begin{subfigure}{0.5\textwidth}
        \includegraphics[width=\linewidth]{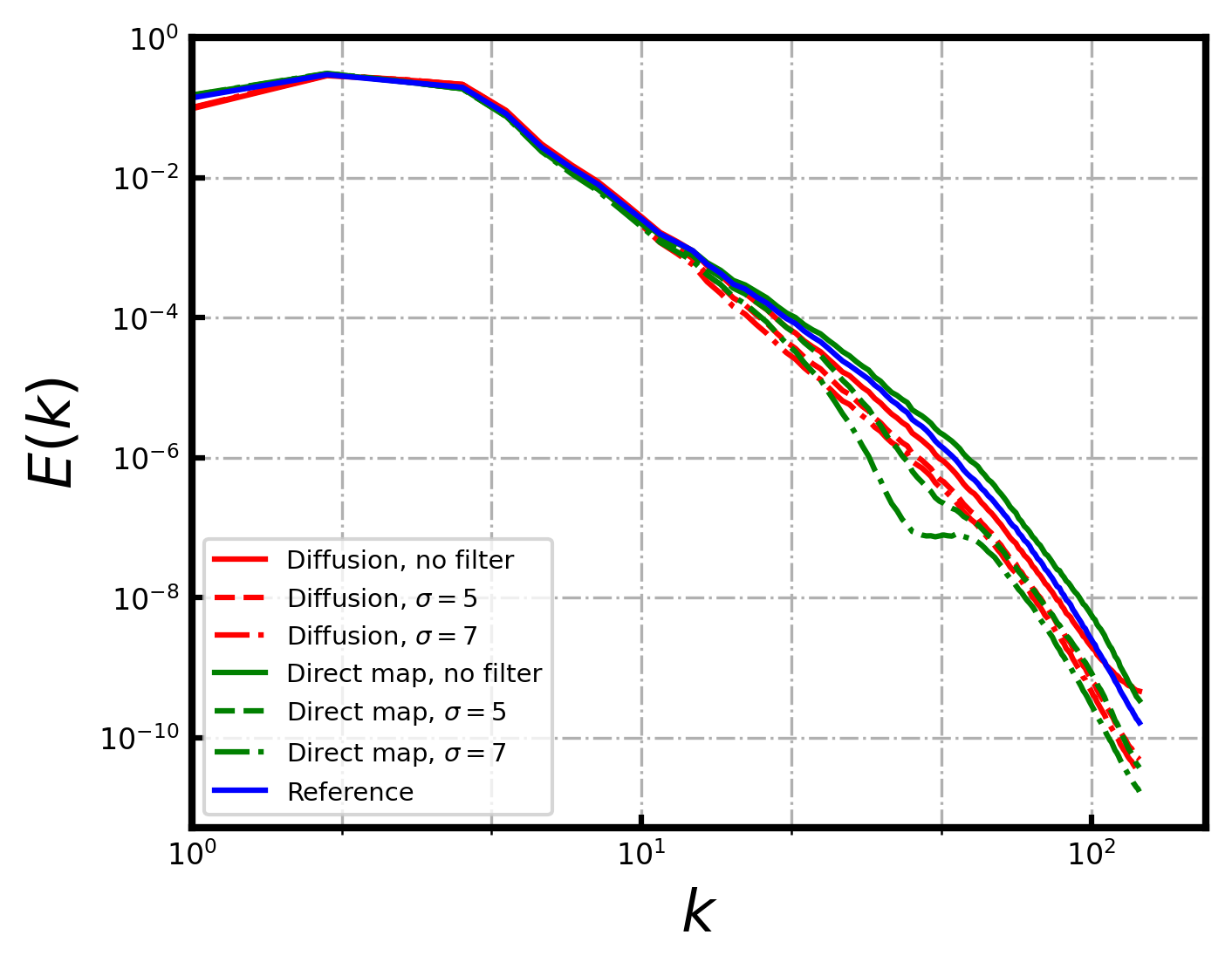}
        \captionof{figure}{Energy spectrum of different models' prediction on sparse reconstruction task.\label{vis-dm-vs-diff-energy}}
    \end{subfigure}
    \caption{Comparison of the UNet that learns a direct mapping and the UNet that learns to estimate noise in the diffusion process.}
\end{figure}
\begin{table}
    \centering
    \begin{tabular}{c|cc|cc} 
    \toprule
    \multirow{2}{*}{Data} & \multicolumn{2}{c|}{$L_{2}$ norm} & \multicolumn{2}{c}{Equation loss}  \\ 
    \cmidrule{2-5}
                      & Diffusion & Direct map            & Diffusion & Direct map             \\ 
    \midrule
    No filter             & 1.8802    & 2.1144                & 0.5428    & 42.9115                \\
    $\sigma=5$            & 1.8213    & 1.9149                & 0.2290    & 17.5249                \\
    $\sigma=7$            & 1.8188    & 1.8408                & 0.1682    & 12.5251                \\
    \bottomrule
    \end{tabular}
    \caption{Quantitative comparison of diffusion model and direct mapping model on sparse reconstruction task with $1.5625\%$ sparsity using different filter scales. \label{tab2-quant-filter}}
\end{table}

\begin{figure}[H]
    \begin{subfigure}{0.49\textwidth}
        \includegraphics[width=\linewidth]{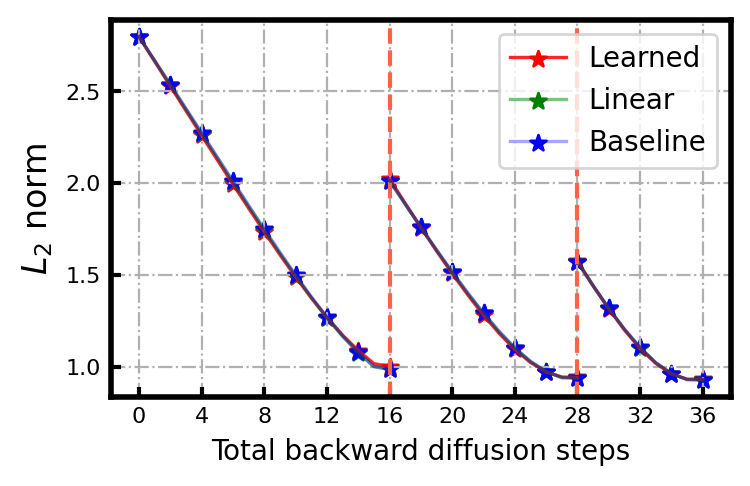}
        \captionof{figure}{Number of steps - $L_2$ norm \label{fig-abla-l2-1}}
    \end{subfigure}
    \begin{subfigure}{0.5\textwidth}
        \includegraphics[width=\linewidth]{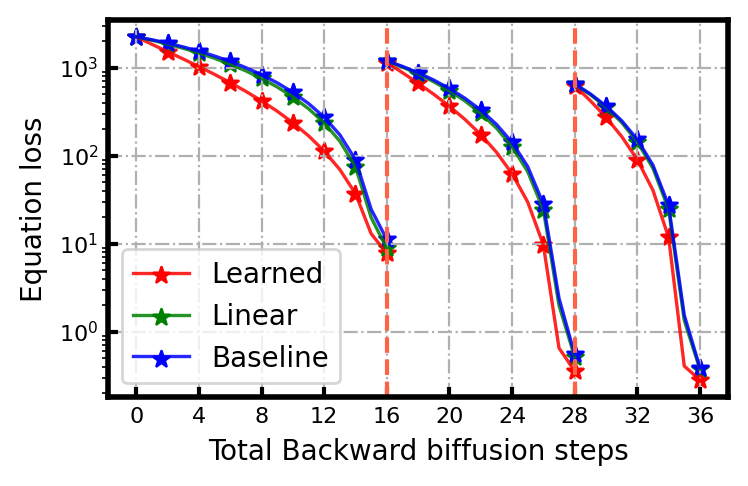}
        \captionof{figure}{Number of steps - Equation loss (log scale)\label{fig-abla-res-1}}
    \end{subfigure}
    \caption{Convergence plots on sparse (5\% collocation points) reconstruction using different methods of combining residual information. The vertical red dotted line indicates a new Gaussian noise injection. }
\end{figure}

\begin{figure}[H]
    \begin{subfigure}{0.49\textwidth}
        \includegraphics[width=\linewidth]{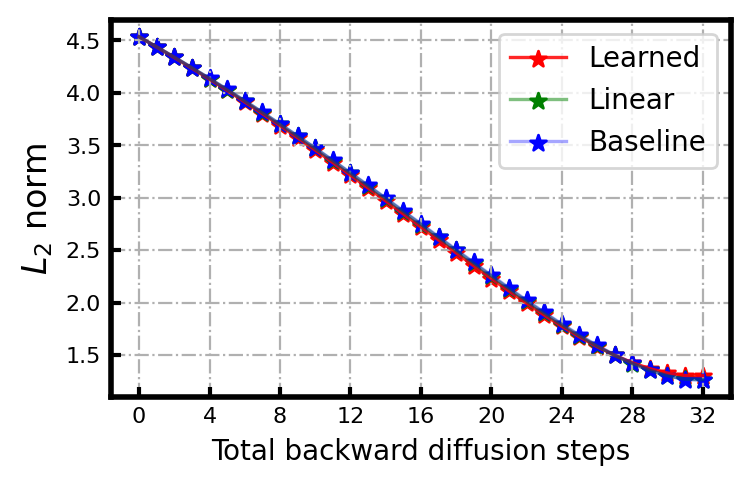}
        \captionof{figure}{Number of steps - $L_2$ norm \label{fig-abla-l2-2}}
    \end{subfigure}
    \begin{subfigure}{0.5\textwidth}
        \includegraphics[width=\linewidth]{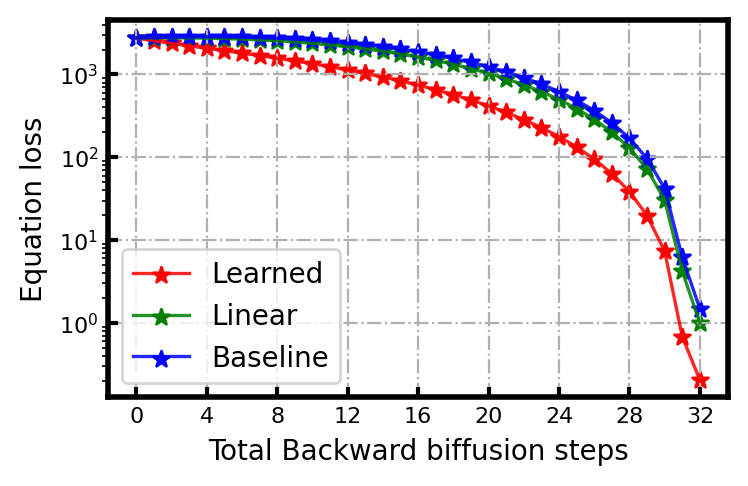}
        \captionof{figure}{Number of steps - Equation loss (log scale)\label{fig-abla-res-2}}
    \end{subfigure}
    \caption{Convergence plots on $8\times$ upsampling using different methods of combining residual information. The vertical red dotted line indicates a new Gaussian noise injection. }
\end{figure}

We also study the influence of adding physics-informed guidance using different ways. The two different ways are the learned and linear combination between the gradient of data distribution and the gradient of residuals. First, all models have a similar trend on $L_{2}$ norm and a negligible difference in the final results (Figure \ref{fig-abla-l2-1}, \ref{fig-abla-l2-2}). Yet with the residual information, diffusion model converges faster on the equation loss and has a lower final residual (Figure \ref{fig-abla-res-1}, \ref{fig-abla-res-2}, especially for more challenging task where stronger Gaussian noise are added to the input data. More specifically, we observe better convergence for the learned combination which combines the residual gradient with the score function non-linearly via the neural network. This is because at the early stage of backward diffusion where the sampled data is very noisy and thus the residual gradient is not very informative, so balancing score function with the gradient of density distribution using a learned neural network is often better than simply combining them linearly.

\begin{table}[H]
\centering
\begin{tabular}{c|cccc!{\vrule width \lightrulewidth}cccc} 
\toprule
\multirow{3}{*}{Method} & \multicolumn{4}{c|}{$L_{2}$ norm} & \multicolumn{4}{c}{Equation loss} \\ 
\cmidrule{2-9}
 & Iter 1 & Iter 1 & Iter 2 & Iter 3 & Iter 1 & Iter 1 & Iter 2 & Iter 3 \\
 & $S=8$ & $S=16$ & $S=30$ & $S=36$ & $S=8$ & $S=16$ & $S=24$ & $S=36$ \\ 
\midrule
Learned & 1.74 & 1.00 & 0.94 & 0.93 & 419.80 & 7.79 & 0.36 & 0.27 \\
Linear & 1.75 & 0.99 & 0.94 & 0.93 & 755.23 & 8.83 & 0.52 & 0.38 \\
Baseline & 1.75 & 0.99 & 0.94 & 0.93 & 832.42 & 11.17 & 0.55 & 0.39 \\
\bottomrule
\end{tabular}
\vspace{+5mm}
\begin{tabular}{c|cccc!{\vrule width \lightrulewidth}cccc} 
\toprule
\multirow{3}{*}{Method} & \multicolumn{4}{c|}{$L_{2}$ norm} & \multicolumn{4}{c}{Equation loss} \\ 
\cmidrule{2-9}
 & Iter 1 & Iter 1 & Iter 1 & Iter 1 & Iter 1 & Iter 1 & Iter 1 & Iter 1 \\

 & $S=8$ & $S=16$ & $S=24$ & $S=32$ & $S=8$ & $S=16$ & $S=24$ & $S=32$ \\ 
\midrule
Learned & 3.69 & 2.72 & 1.77 & 1.30 & 1572.02 & 744.18 & 177.68 & 0.20 \\
Linear & 3.70 & 2.74 & 1.79 & 1.26 & 2546.79 & 1623.54 & 498.23 & 0.99 \\
Baseline & 3.70 & 2.75 & 1.80 & 1.26 & 2740.99 & 1876.36 & 621.12 & 1.47 \\
\bottomrule
\end{tabular}
\caption{A comparison of convergence trend for different methods of combining residual information, where the baseline is the original diffusion model that does not use any residual information. $S$ denotes the total number of backward diffusion steps, while \textit{Iter} denotes the outer diffusion loop as described in Algorithm \ref{alg:ddpm-conditional-sampling-conditional}. The upper table are evaluated on the non-equidistant sparse reconstruction task, the bottom table are evaluated on the $8\times$ upsampling task.}
\end{table}

As shown in the previous part of Results section, we have selected three performance metrics to evaluate the model for high-fidelity CFD data reconstruction: 1) $L_2$ norm, 2) equation loss, and 3) kinetic energy spectrum. $L_2$ norm is a straightforward way to measure the reconstruction error with respect to the ground truth and has been widely used as a performance metric in many data prediction tasks. $L_2$ norm is also the loss function used in the training and test of the direct mapping method that benchmarks our proposed method. However, the $L_2$ norm has two limitations in properly evaluating the CFD data reconstruction. First, the $L_2$ norm is less sensitive to the blurring of data. In the case where a large amount of blurring effect is added (\textit{e.g.}, through Gaussian blur) to a CFD data sample, its $L_2$ norm tends to have an insignificant increase, which is counter-intuitive to a human observer. Second, despite being a popular universal metric to measure the distance between two data points, $L_2$ norm is not specialized to reveal how accurately a model's prediction reflects the physical characteristics of the fluid, for example, it does not indicate how well the prediction fits the Navier-Stokes equation, or whether the eddies of different scales in the prediction contain the accurate amount of turbulence kinetic energy. 
While the $L_2$ norm is a direct indicator of the averaged point-wise reconstruction error, other metrics such as the equation loss and the kinetic energy spectrum help to reveal how well a reconstructed CFD data sample fits the expected physical characteristics seen in the ground truth sample. Therefore, we include equation loss and kinetic energy spectrum in addition to $L_2$ norm as the performance metrics in order to have a more comprehensive evaluation of the reconstruction methods.
We consider that our proposed method outperforms the benchmark methods such as the direct mapping model and the bicubic interpolation, given the case that our model shows a marginal improvement on $L_2$-loss, a significant improvement on equation loss, and a closer alignment to the ground truth in terms of the kinetic energy spectrum, as such experimental result indicates that the reconstruction obtained by our proposed method has a comparable $L_2$-loss accuracy with the benchmark method while being more coherent with the ground truth physical characteristics of the fluid.

\section{Conclusion}
In this work, we presented a diffusion model for high-fidelity CFD data reconstruction from low-fidelity input. We showed that a diffusion model is able to solve the data reconstruction problem as a conditional data denoising problem. Compared with the benchmark method which learns the direct-mapping from low-fidelity to high-fidelity data, our model has a similar (marginally better) reconstruction accuracy in terms of the $L_2$ loss, while having the advantage of being much more robust to variation in the low-fidelity input data and being more accurate in terms of data kinetic energy spectrum. In addition, we showed how to incorporate the physics-informed information such as the PDE residual gradient in model training and model inference for an improved performance, and how to reconstruct high-fidelity CFD data from sparsely measured inputs using nearest padding and iterative reconstruction.

Compared with the direct-mapping models, a diffusion model is not trained to directly minimize the reconstruction loss in the sense of an $L_{p}$ norm. Instead, it is trained to minimize the KL-divergence between the forward diffusion process and the backward diffusion process. As a result, a diffusion model is essentially designed to be only sensitive to data reconstruction error in a statistical sense with the presence of Gaussian noise. We consider the absence of an $L_{p}$ reconstruction loss as a potential limitation that prevents a diffusion model from further improving its reconstruction accuracy. To resolve such limitation, a new design of the data sampling procedure for the diffusion model is needed. Alternatively, it might be interesting to investigate an ensemble learning method which incorporates a diffusion model and a direct-mapping model, so that the merits of both methods can be retained. These potential ways to resolve the limitation of diffusion models for data reconstruction will be the main direction of our future work.

\section{Acknowledgement}
This work is supported by the start-up fund from the Department of Mechanical Engineering, Carnegie Mellon University, United States.

\newpage
\bibliography{ref}
\end{document}